# Estimation of Gas Turbine Shaft Torque and Fuel Flow of a CODLAG Propulsion System Using Genetic Programming Algorithm

Nikola Anđelić, Sandi Baressi Šegota, Ivan Lorencin, Zlatan Car
Faculty of Engineering, University of Rijeka, Vukovarska 58, 51000 Rijeka, Croatia
Email: nandelic@riteh.hr, sbaressisegota@riteh.hr, ilorencin@riteh.hr, zlatan.car@riteh.hr

In this paper, the publicly available dataset of condition based maintenance of combined diesel-electric and gas (CODLAG) propulsion system for ships has been utilized to obtain symbolic expressions which could estimate gas turbine shaft torque and fuel flow using genetic programming (GP) algorithm. The entire dataset consists of 11934 samples that was divided into training and testing portions of dataset in an 80:20 ratio. The training dataset used to train the GP algorithm to obtain symbolic expressions for gas turbine shaft torque and fuel flow estimation consisted of 9548 samples. The best symbolic expressions obtained for gas turbine shaft torque and fuel flow estimation were obtained based on their $R^2$ score generated as a result of the application of the testing portion of the dataset on the aforementioned symbolic expressions. The testing portion of the dataset consisted of 2386 samples. The three best symbolic expressions obtained for gas turbine shaft torque estimation generated $R^2$ scores of 0.999201, 0.999296, and 0.999374, respectively. The three best symbolic expressions obtained for fuel flow estimation generated $R^2$ scores of 0.995495, 0.996465, and 0.996487, respectively.

## 1 Introduction

So far, the standard approach in maintenance, according to [1], was to fix it when it breaks. However, in the last few decades and due to the high repairing costs, the smart technologies and cross-industry needs in maintenance have caused a shift change from a reactive to a proactive perspective. Instead of classical repairing-replacing maintenance actions the industry, in general, has focused more on preventive-perspective activities. The maintenance actions, according to [2], can be divided into three categories, and these are:
- corrective maintenance - this type of maintenance is triggered by unscheduled events for example system failure. Before the system failure, no prior maintenance action strategies, are applied. This maintenance approach is very expensive when compared to the other two, due to direct (concatenated failures of other system parts) and indirect costs related to potential losses in safety and integrity and asset unavailability,
- preventive maintenance - this type of maintenance is carried out before system breakdowns to avoid or minimize potential issues related to failures. There are several variations of preventive maintenance such as adjustments, replacements, renewals, and inspections. Compared to the previous maintenance type this type allows to establish time-slots of

- unavailability, and
- condition-based maintenance (CBM) - type of maintenance in which maintenance activities are activated based on the condition of the target system. This approach enables determining the condition of system elements that can be used in order to predict the potential degradation and to consequently plan when maintenance activities will be required and performed which can result in system disruption minimization. The CBM have switched the maintenance view from pure diagnosis to the high-valued prognosis of faults.

The CBM is usually used in marine propulsion systems or its components since it enables a just-in-time deployment of ship maintenance, by allowing to plan and execute maintenance activities when needed. This maintenance approach requires a reliable and effective diagnostic policy on one or more naval asset, in order to further refinement toward identifying potential failures in advance. According to [3, 4] the gas turbines for naval propulsion represents a key example of how gathering data from components can be a used to optimize maintenance strategies by utilizing diagnostic methods that allow to obtain useful information on the status of components without the need to inspect the machine. In order to develop prognostic models for GTs the diagnostic is the starting point that is used to describe its status.

The marine propulsion systems are used to generate thrust to move ship or boat across water. Modern ships today are propelled by different power sources such as steam turbines [5 - 10] (nuclear-powered steam turbines [11, 12]), turbo-electric transmission [13], diesel [14, 15], and reciprocating diesel engines [16], LNG engines [17], gas turbines [18], Stirling engines [19, 20], etc. Many warships at the beginning of the second half of the last century have used gas turbines for propulsion. Due to the low thermal efficiency of gas turbines at low power output, diesel engines have been used for cruising while gas turbines used for higher speeds. Another important factor according to [21] for utilization of gas turbines was to reduce emissions in sensitive environmental areas or in ports. Another alternative is to combine steam turbines to improve efficiency of their gas turbines in a combined cycle. The reason for a such combination is that the waste heat from gas turbine exhaust is utilized to boil water and create steam for driving the steam turbine. All these combination of different propulsion systems together created combined marine propulsion system that have some advantages when compared to singular propulsion system.

The combined diesel-electric and gas (CODLAG) is a variant of combined diesel and gas propulsion system for ships. This system according to [22 - 27] employs electric motors which are connected to the propeller shafts and these motors are powered by diesel generators. In order to achieve higher speeds, a gas turbine powers the shafts using a cross-connecting gearbox, and for cruise speed, the turbine drive train is disengaged with clutches. There are a few papers that describe the investigation of such a complex system. In [25], the authors have simulated the dynamic behavior of the CODLAG propulsion plant during transients and off-design conditions. The methodology and the simulation models that are needed to design the propulsion control logic of the CODLAG propulsion plant have been investigated in [28]. The simulation study of performance characteristics of the CODLAG propulsion system has been investigated in [29]. In the majority of scientific papers, the investigation of the control system and vehicle hydrodynamics simulation is very detailed, while the propulsion system is often absent or very simplified. However, there are a few papers that investigated the propulsion plant topics, i.e. gas turbines [30 - 35] and diesel engines [36 - 43]. The mechanics, hydraulic systems, and acting loads of propeller pitch change mechanisms have been investigated in [44 - 47].

Today the artificial intelligence (AI) is implemented in various fields such as energy sector [48 - 51], medicine [52 - 55], maritime [56 - 59], economics [60] and etc. There are some research papers in which the AI algorithms have been implemented on the CODLAG dataset to perform some estimation of certain parameters. In [61], the authors have implemented Artificial Neural Network (ANN) and ANN with Principle Component Analysis (PCA) to model and predict turbocompressor decay state coefficient and turbine decay state coefficient of a GT mounted on a frigate characterized by CODLAG propulsion plant used in naval vessels. The multi-layer perceptron (MLP) was applied, on CODLAG publically available dataset in [26] to predict the gas turbine (GT) and GT turbocompressor decay state coefficient. In [27] the same dataset was used for MLP training and testing to estimate the frigate speed of the CODLAG propulsion system. The propeller torques of a frigate using the CODLAG dataset were estimated using ANN, as reported in [62].

The genetic programming algorithm can be described as an AI technique of evolving programs that start from a population of unfit, randomly generated programs and improving their fitness from generation through generation with the application of genetic operators (crossover and mutation). In each generation, the selection of the fittest programs (programs with higher fitness value) is performed, which are used, for reproduction (crossover operator) and mutation. However, the key factor in this algorithm, as in the genetic algorithm (GA), is the proper definition of the fitness measure. The application of crossover operation requires two programs (parents) and in this operation, the genetic material is randomly selected and exchanged between them to produce the offspring (children) that become part of the next generation of programs. Mutation operation, on the other hand, requires only one population member. The genetic material is randomly selected on population members and substituted with randomly generated genetic material. In the majority of cases, the members of the next generation are more fitted than members of the previous generation.

The history of evolving programs dates back to Alan Turing [63] considered as the father of AI. However, due to the computational limitation of that time, there is a gap of almost 30 years before the successful evolution of small programs was achieved [64]. The invention of GA for program evolution was patented by Koza in 1988 [65] which was later followed by publication in International Joint Conference on Artificial Intelligence [66]. Koza published over 200 publications and 4 books about Genetic Programming, and by doing so, the field of GP was established. Today the GP was applied to various fields, such as:
- curve fitting, data modeling and symbolic regression [67 - 69],
- image and signal processing [70 - 72],
- financial trading, time series prediction and economic modeling [73 - 75],
- industrial process control [76 - 78], and
- medicine, biology and bioinformatics [79 - 81].

Although there are a number of scientific papers in which GP has been successfully applied the application of this algorithm is very rare when compared to other AI methods such as ANN, GA, etc. When solving a specific problem using ANN the model of ANN is trained and later tested to solve a specific problem. However, the ANN model could not be translated into a mathematical equation due to a large number of interconnected neurons. The methodology in GP is similar to ANN since the GP algorithm requires a dataset that is divided into training and testing datasets. The training portion of the dataset is used in the GP algorithm to obtain symbolic expression which correlates the input values with the desired output value. The testing portion of the dataset is used to test the obtained symbolic expression on the unseen part of the dataset and to validate the obtained symbolic expression. When compared to the ANN the benefit of the GP algorithm is that the result of training the GP algorithm with a specific dataset is the mathematical expression (symbolic expression) which correlates input values with the desired output with the help of mathematical functions and constant values. Based on previously detailed literature overview and comparison of GP algorithm with other AI methods the following questions arise and these are:
- Is it possible to utilize the GP algorithm on the CODLAG dataset in order to obtain symbolic expressions that could estimate the gas turbine shaft torque with the influence of gas turbine turbocompressor decay and gas turbine decay state coefficients?
- Is it possible to utilize the GP algorithm in order to obtain symbolic expressions that could estimate fuel flow with the influence of gas turbine turbocompressor decay and gas turbine decay state coefficients?

## 2 Materials and Methods

In this section the detailed overview of the publicly available CODLAG dataset used in GP algorithm, and the description of GP algorithm is given.

### 2.1 Dataset Description

The CODLAG dataset used to obtain symbolic expressions of gas turbine shaft torque and fuel flow estimation using GP algorithm is a publicly available dataset from UCI Machine Learning repository [22]. In order to obtain the dataset authors [22 - 24] have carried out the experiments by means of a numerical simulator of a naval vessel which is characterized by gas turbine propulsion plant. This numerical simulator of naval vessel with gas turbine propulsion plant consist of blocks such as: propeller, hull, gas turbine, gear box and controller that were developed and fine tuned on several similar real propulsion plants. The electrical power required for the operation of electric motors is powered with diesel-generators (diesel engine that powers electrical generators). The Firgate propeller is driven from power generated with GT and two electrical motors which is transmitted using a system that consists of three gear boxes and four clutches. The schematic view of the CODLAG propulsion system is shown in Fig. 1.

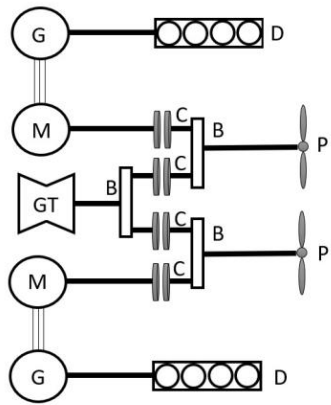

Figure 1. The schematic view of CODLAG propulsion system [26]
(GT - gas turbine; M - electric motor; G - electric generator; D - diesel engine; B - gear box; C -

clutch; P - Frigate propeller)

The GT of the CODLAG propulsion system consist of turbocompressor, combustion chamber, high pressure (HP) and low pressure (LP) gas turbines. The power produced in HP GT is utilized only for turbocompressor drive and the power produced with LP GT is used for ship propulsion, together with power produced by electric motors. The schematic view of the GT used in CODLAG propulsion system is shown in Fig. 2.

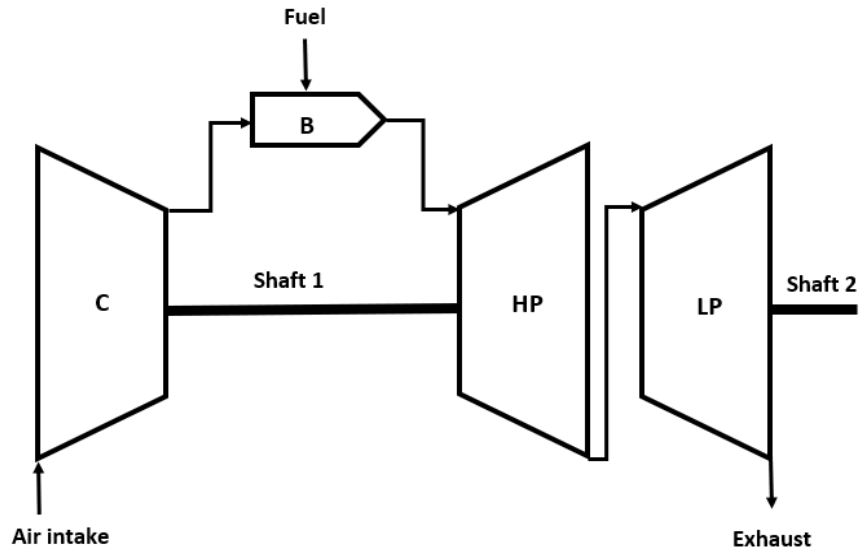

Figure 2. The schematic view of gas turbine used in CODLAG propulsion system [26]

As seen from Fig. 2 the HP GT, in this configuration and along with turbocompressor and combustion chamber, is utilized as the gas generator. The LP GT has no mechanical connection with the HT GT. However, the LP GT receives the flue gas from the HP GT which is used to produce power for shaft turbine.

The behavior of the propulsion system is described with parameters such as: ship speed, turbocompressor degradation coefficient and turbine degradation coefficient. This means that each possible degradation state can be described by a combination of previously mentioned parameters. The simulator as well as dataset takes into the account a performance decay over time of GT components such as GT turbocompressor and turbines. These decay coefficients (turbocompressor and turbine) have been created using uniform grid with precision of 0.001 in order to achieve good representation. Based on conducted simulations the dataset containing 18 features (16 features with 2 decay coefficients) with 11934 instances has been obtained. In Tab. 1 the parameters of dataset with corresponding range of values and units is given.

Table 1. The list of physical values in CODLAG dataset with corresponding range of values and units.

| Physical variable | Range | Unit |
|---|---|---|
| Lever position ($l_p$) | 1.138-9.3 | - |
| Ship speed ($v$) | 3-27 | kn |
| Gas turbine shaft torque (GTT) | 253.547-72784.872 | kNm |
| GT rate of revolutions (GTn) | 1307.675-3560.741 | rpm |
| Gas generator rate of revolutions (GGn) | 6589.002-9797.103 | rpm |
| Starboard propeller torque (Ts) | 5.304-645.249 | kN |
| Port propeller torque (Tp) | 5.304-645.249 | kN |
| High pressure turbine exit temperature (T48) | 442.364-1115.797 | °C |
| GT turbocompressor inlet air temperature (T1) | 288 | °C |
| GT turbocompressor outlet air temperature (T2) | 540.442-789.094 | °C |
| HP turbine exit pressure (P48) | 1.093-4.56 | bar |
| GT turbocompressor inlet air pressure (P1) | 0.998 | bar |
| GT turbocompressor outlet air pressure (P2) | 5.828-23.14 | bar |
| GT exhaust gas pressure ($P_{exh}$) | 1.019-1.052 | bar |
| Turbine injection control (TIC) | 0-92.556 | % |
| Fuel flow ($m_f$) | 0.068-1.832 | kg/s |
| GT turbocompressor decay state coefficient | 0.95-1 | - |
| GT turbine decay state coefficient | 0.975-1 | - |

**Source:** Authors

## 2.2 GP algorithm

GP algorithm is unique AI method used to obtain symbolic expressions from the dataset with highest correlation between input and output values. The algorithm can be described as the combination of machine learning and evolutionary computation. The similarities of GP with machine learning is that GP requires the dataset which is then divided into training and testing dataset. The same methodology is used in machine learning methods i.e. supervised learning method. Using the training dataset GP algorithm generates the symbolic expression which creates correlation between input and output parameters. After the symbolic expression is obtained on training dataset, the testing dataset is used to validate obtained symbolic expression. GP algorithm have some similarities as evolutionary algorithms in terms of population, fitness function and crossover and mutation operators. After creation of the initial population of symbolic expressions these expressions are then evaluated using fitness function in order to determine better symbolic expressions which will then represent the parents of next generation. On better population members the crossover and mutation operations are performed in order to obtain offspring which will be a part of next generations on symbolic expressions. These operations are performed from generation to generation until some stopping criteria was met.

### 2.2.1 Basic Structure of GP Algorithm

The basic structure of GP Algorithm consists of:
- population initialization,
- evaluation (fitness function),
- selection,
- variation operators (crossover and mutation), and
- stopping criteria.

### 2.2.2 Population Initialization

In order to describe the method used to generate the initial population the representation of population members must be explained first. GP has unique representation of each population member since they are expressed as syntax trees. The tree representation of symbolic expression $(X_2 - X_7) + (1 - X_1)$ is shown in Fig. 3.

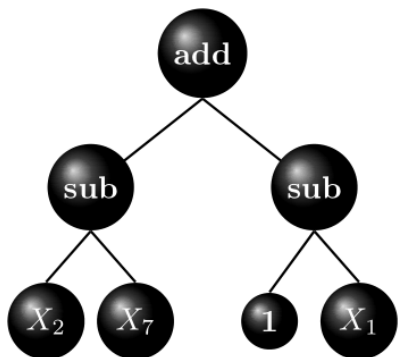

Figure 3. Example of GP syntax tree representing the symbolic expression $(X_2 - X_7) + (1 - X_1)$ **Source:** Authors

As seen from Fig. 3 the variables ($X_1, X_2$ and $X_7$) and constant (1) in the syntax tree are represented as leaves of the tree while arithmetic operations (addition (add), and subtraction (sub)) are represented as internal nodes of the syntax tree. In GP the variables and constants are chosen from the terminal set while the arithmetic operations are chosen from function set. These two sets together form primitive set of a GP system. In GP there are two subsets of primitive set and these are terminal and function set which are used to create population members. The terminal set usually consists of:
- the symbolic expression external inputs - input and output values from the training dataset which are represented as symbolic expression variables $(x_i, y)$,
- the mathematical functions without arguments - these functions are included in the terminal set because each time they are used they return different values. Example of these function type is the function rand() which returns random numbers, and
- the constants - are defined in prespecified range, and they are randomly generated as part of the initial tree creation process, or created by mutation operator.

In GP the function set is usually defined based on the complexity of the problem. If a problem which is solved with GP algorithm is simple numeric problem it is usually recommended that the function set consists of simple arithmetic functions (addition, subtraction, multiplication and division). However, if the problem is far more complex than the definition and implementation of advanced mathematical functions must be considered. In GP as in GA the population members in the initial population are usually randomly generated. Over the years the various methods have been developed for population initialization. However, in this paper the ramped half-and-half method was utilized. This method is a combination of two earliest population initialization methods and these are full and grow method. In both methods the population members are generated in the way so they do not exceed the prespeified maximum depth. In syntax tree structure the depth of a node is the number of edges that need to be traversed to reach the node starting from the root node. The full method was named in that way since the syntax tress generated by this method are full tress or in other words the leaves are at the same depth. When syntax tree structures are created using full method the nodes are randomly chosen from function set until the maximum tree depth is reached. After maximum depth of syntax tree that consists of functions are chosen the leaves are selected from the terminal set (i.e variables and constants). Using full method the generated initial population members have leaves at the same depth. However, that does not mean that all initial trees will have same number of nodes.

The grow method, when compared to full method, will generate syntax trees with varied sizes and shapes. Unlike full method, in grow method, the nodes are selected from entire primitive set until the depth limit is reached. This means that nodes can be functions, constants or variables. Once the predefined tree depth is reached only terminals may be chosen.

When compared both methods do not provide a wide variety of syntax tree sizes and shapes so ramped half-and-half was developed as combination of full and grow method. Using ramped half-and-half method, half of the initial population is constructed using full and half is constructed using grow method. However, the members of initial population do not have equal depth because the tree depth limit is prespecified in certain range.

### 2.2.3 Fitness Function

The fitness function is one of the key elements of GA as well as GP. The fitness function can be described as a quality measure of population member obtained as a results of its evaluation. There are different types of fitness functions that can be used in GP and these are:
- the amount of error between its output and the desired output,
- the amount of time needed for system to reach target state,
- the accuracy of population member in recognizing patterns or classifying objects, and
- the payoff that a game-playing program produces.

In all GP simulations the mean absolute error (*MAE*) was utilized as a fitness function which can be written in the following form

$$MAE = \frac{\sum_{i=1}^{n}|y_i - x_i|}{n}, \qquad (1)$$

where $y_i$ is prediction, $x_i$ is the true value and $n$ is the number of instances. However, after the symbolic expression is obtained in each GP run using training portion of the dataset the symbolic expression is than evaluated with coefficient of determination ($R^2$) using the testing portion of the dataset. The $R^2$ metric of each symbolic expression is calculated using expression which can be written in the following form

$$R^2 = 1 - \frac{S_{RESIDUAL}}{S_{TOTAL}} = 1 - \frac{\sum_{i=0}^{m}(y_i - \hat{y}_i)^2}{\sum_{i=0}^{m}(y_i - \frac{1}{m}\sum_{i=0}^{m} y_i)^2}. \tag{2}$$

This metric compares two set of solutions, in terms of variance, and these are the real data $y$ and the data obtained by the model $\hat{y}$. In other words, the $R^2$ calculates the amount of variance contained insed the data $y$, which is explained in the data $\hat{y}$ as a model output. The $R^2$ result is the value which is in range from 0 to 1. The $R^2$ value of 1.0 means that there is no variance between the real data and the data obtained by the model, and the $R^2$ value of 0.0 means none of the variance in the real data is explained in the model data.

### 2.2.4 Selection

As in other EA, in GP the genetic operators are applied on the population members that are selected based on their quality measure obtained with fitness function evaluation with certain probability. In other words, better individuals will have more offspring's than inferior individuals. Today there are numerous methods that are used for selection of population members. However, in GP the tournament selection will be utilized.
The tournament selection is method of randomly selecting a number of individuals from the population. These randomly selected population members are then compared with each other and the best population members are chosen as parents.

### 2.2.5 Genetic Operations

The usual, well known, genetic operations used to create members of next generation are crossover and mutation. Crossover requires at least two population members to generate offspring. From these two parents randomly selected genetic material is used to form the member of the next generation. The mutation operator requires one population member. On this population member the randomly selected part is switched with randomly generated part created using primitive set. Since in GP each population member is represented as tree structure there are three different mutation operators which can be used and these are: subtree mutation, hoist mutation and point mutation. Subtree mutation is mutation operator that takes the winner of a tournament selection and selects the random subtree (branch) that will be replaced with the donor subtree. The donor subtree is randomly generated using the primitive set and it is inserted into the parent to form an offspring of next generation. Hoist mutation is the mutation operator that takes a winner from tournament selection and randomly selects subtree for mutation. After random subtree selection the random subtree of that subtree is selected and is hoisted into the original subtrees location in order to form offspring which will participate in next generation. The point mutation is the mutation operator that takes the winner of tournament selection and randomly selects the tree node which will be replaced. It should be noted that terminal nodes are replaced with other randomly selected terminal, and functions are replaced by other randomly selected function. However, the function is replaced by other function that require the same number of arguments as the original node. After point mutation operation is performed the offspring is set to be a part of next generation.

### 2.2.6 Stopping Criteria

In GP as in other EA the stopping criteria is the most crucial part of the entire algorithm. This part is responsible for stopping GP execution after certain criteria was met. In GP there are two main stopping criteria that are usually used to stop the execution of each simulation and these are number of generations and the fitness value. The number of generation parameter is user defined value which is responsible for stopping the GP algorithm execution after maximum number of generations is reached. The fitness value is the other parameter defined by the user which stops the execution after the fitness value generated with one population member is reached. In this paper the fitness value (*MAE*) was never reached so the maximum number of generations was utilized as the alternative stopping criteria.

## 3   Results and Discussion

For better representation the input and output variables with corresponding labels are given in Tab. 2.

Table 2. The physical variables of the dataset with corresponding variable type used in GP algorithm to obtain gas turbine shaft torque and fuel flow symbolic expressions. $X_i$ represent input variables while $y$ represent output variable.

| Physical Variable | Representation of variables in GP algorithm | |
|---|---|---|
| | Gas Turbine Shaft Torque Analysis | Fuel Flow Analysis |
| Lever Position ($l_p$) | $X_0$ | $X_0$ |
| Ship Speed ($v$) | $X_1$ | $X_1$ |
| Gas turbine shaft torque | $y$ | $X_2$ |
| GT rate of revolutions (GTn) | $X_2$ | $X_3$ |
| Gas generator rate of revolutions (GGn) | $X_3$ | $X_4$ |
| Starboard propeller torque ($T_s$) | $X_4$ | $X_5$ |
| Port propeller torque ($T_p$) | $X_5$ | $X_6$ |
| High pressure turbine exit temperature ($T_{48}$) | $X_6$ | $X_7$ |
| GT turbocompressor inlet air temperature ($T_1$) | $X_7$ | $X_8$ |
| GT turbocompressor outlet air temperature ($T_2$) | $X_8$ | $X_9$ |
| HP turbine exit pressure ($P_{48}$) | $X_9$ | $X_{10}$ |
| GT turbocompressor inlet air pressure ($P_1$) | $X_{10}$ | $X_{11}$ |
| GT turbocompressor outlet air pressure ($P_2$) | $X_{11}$ | $X_{12}$ |
| GT exhaust gas pressure ($P_{exh}$) | $X_{12}$ | $X_{13}$ |
| Turbine injection control (TIC) | $X_{13}$ | $X_{14}$ |
| Fuel Flow ($m_f$) | $X_{14}$ | $y$ |
| GT turbocompressor decay state coefficient | $X_{15}$ | $X_{15}$ |
| GT turbine decay state coefficient | $X_{16}$ | $X_{16}$ |

**Source:** Authors

As seen from Tab. 2 in both analyses there are total of 17 input variables that are indicated with $X_i$ where $i$ is in range from 0 to 16. The output variable in both analyses are indicated with the $y$. The GP parameters used to obtain symbolic expressions for gas turbine shaft torque and fuel flow estimation are different and are described in following subsections. However, the mathematical functions used to create symbolic expressions in both cases are the same and are shown in Tab. 3.

Table 3. The list of functions used from function set in GP algorithm to create symbolic expressions.

| Function set | |
|---|---|
| Kind of primitive | Example of functions |
| Arithmetic | addition (+) |
| | subtraction (-) |
| | multiplication (*) |
| | division (/) |
| Mathematical | square root (sqrt()) |
| | absolute value (abs()) |
| | logarithm (log()) |
| | maximum (max()) |
| | minimum (min()) |
| | sine (sin()) |
| | cosine (cos()) |
| | tangent (tan()) |

**Source:** Authors

## 3.1 Results Obtained for Gas Turbine Shaft Torque

In this case the training and testing portions of the dataset had 17 input variables and gas turbine shaft torque values as output variable. The 17 input variables are labeled from $X_0, \ldots, X_{16}$ while the output variable is labeled as $y$. The physical variables of the dataset with corresponding variable type used in GP algorithm to obtain the symbolic expressions for gas turbine shaft torque estimation are shown in Tab. 2. The parameters range used to obtain symbolic expression for gas turbine shaft torque estimation are shown in Tab. 4.

Table 4. The range of parameters used in GP to obtain symbolic expression for gas turbine shaft torque estimation

| Parameter | Lower Bound | Upper Bound |
|---|---|---|
| Population size | 50 | 100 |
| Number of Generations | 100 | 1000 |
| Tournament size | 5 | 30 |
| Three depth | (3,7) | (6,12) |
| Crossover Coefficient | 0.7 | 1 |
| Subtree mutation coefficient | 0.01 | 0.1 |
| Hoist mutation coefficient | 0.01 | 0.1 |
| Point mutation coefficient | 0.01 | 0.1 |
| Stopping criteria | $1.0 \times 10^{-6}$ | 0.001 |
| Maximum Samples (%) | 70 | 100 |
| Constant range | -100 | 100 |
| Parsimony coefficient | 0.1 | 1 |

**Source:** Authors

The three best symbolic expressions for gas turbine shaft torque estimation with GP parameters and corresponding $R^2$ score are shown in Tab. 5.

Table 5. The best three symbolic expressions obtained for gas turbine shaft torque estimation with GP parameters and $R^2$ score.

| GP Parameters | Symbolic expression | $R^2$ score |
|---|---|---|
| [68, 995, 8, (5, 8), 0.769, 0.0516, 0.0462, 0.096, 0.000434, 0.7795, (-77.601, 44.708), 0.82] | $y_{GTT1} = -327.559 + 2X_1X_4 + (X_1 + X_{11})X_8 +$ $\min\left(-16.592, -932.902 - X_{11} - X_{12} - X_4 + X_1X_4\right.$ $+ \cos(292.2 - X_1(X_4 + X_8)$ $\left.- X_4\sin\left(\frac{-310.9 - X_{11} + X_4\sin(X_5)}{X_{14}}\right)\right.$ $\left.+ \frac{-310.9 + X_{11} + X_4\sin(X_5)}{X_{14}}\right) + \sin(X_5) + X_4(\cos(X_5)$ $+ \sin(X_5))$ | 0.999201 |
| [100, 838, 20, (3, 12), 0.737, 0.097, 0.093, 0.0334, 0.000552, 0.986, (-46.829, 57.09), 0.13] | $y_{GTT2} = \left\|\left(\frac{X_7 + X_8}{X_2X_{14}} - X_{11}\right)X_8\right\| + X_1(X_8 + 2X_4)$ | 0.999296 |
| [74, 947, 23, (5, 7), 0.814, 0.065, 0.049, 0.0629, 0.000662, 0.872, (-3.56, 54.173), 0.3] | $y_{GTT3} = \sqrt{\|A\|}(X_4 + (2B - X_2(C + \sqrt{A}((2B - X_2(C$ $+ \sqrt{C + \sqrt{\sqrt{B} - 2X_3} - X_3)^{\frac{1}{2}})^{\frac{1}{4}}}$ $+ X_4))^{\frac{1}{2}})^{\frac{1}{4}});$ $A = -X_3 + 40.7X_9, B = X_{14}X_2X_3;$ $C = -X_2(X_2 - X_{14}X_3 + \sqrt{X_{14}X_2(X_2 + \sqrt{X_2 - B})})$ | 0.999374 |

**Source:** Authors

As seen from Tab. 5 the best symbolic expressions achieved $R^2$ scores of 0.999201, 0.999296 and 0.999374, respectively. These symbolic expressions were achieved with randomly selected population sizes of 68, 100 and 74, respectively. These population members were evolved through 995, 838 and 947 generations, respectively. When compared to the number of generations shown in Tab. 4 it can be noticed that population size and number of generations value were close to the upper bound of the predefined range. When tournament size values compared in those three cases it can be noticed that the lowest tournament size value was in the case of first symbolic expression (8) while in remaining two cases this value was much higher (20 and 23). The range of maximum tree sizes used for creation of initial population were (5,8), (3,12) and (5,7). From these values it can be noticed that the highest range and the highest maximum tree depth number was in the second case from 3 to 12. The crossover coefficient value was randomly selected in all three cases and the values are 0.796, 0.737, and 0.814. In all three cases the values of the crossover coefficient were bear the lower bound of the predefined range for this coefficient as shown in Tab. 4. The subtree mutation, hoist mutation and point mutation coefficient values in first case are equal to 0.0516, 0.0462, and 0.096; in second case 0.097, 0.093, and 0.0334 while in the third case are equal to 0.065, 0.049 and 0.0629. All these values are low compared to crossover coefficient which means that the dominating genetic operator in all these cases is crossover. The stopping criteria values in all three cases are equal to 0.000434, 0.000552 and 0.000662, respectively. The stopping criteria coefficient is one of the stopping criteria that is responsible for stopping the GP execution if the fitness value of one population member drops below this value. In all these investigations the MAE criteria was used as a fitness measure. However, the MAE value in all three cases did not drop below the stopping criteria coefficient value so maximum number of generations was used as the stopping criteria in each GP analysis. After the GP reached predefined value of maximum number of generations the GP execution was terminated. Maximum number of samples is the fraction of samples from training dataset that are used to evaluate each population member from generation to generation. The values of maximum number of samples were randomly chosen values from predefined range given in Tab. 4 and are equal to 0.7795, 0.986 and 0.872. This means that in the first case only 77.95 % of training dataset was used to evaluate population members, in the second case 98.6 % and in the third case only 87.2 % was used to evaluate population members. The constant range used for creation of population members was randomly created range from predefined range and are equal to (-77.601, 44.708), (-46.829, 57.09) and (-3.56, 54.173), respectively. The parsimony coefficient values in all three cases are equal to 0.82, 0.13 and 0.3. This coefficient is responsible for penalizing large population members in each generation by adjusting their fitness to be less favorable for selection.

When all these equations in Tab. 5 are compared it can be seen that the smallest symbolic expression used for GT shaft torque estimation is the second symbolic expression with $R^2$ of 0.999296. It can be noticed that in first and third symbolic expression the bloat phenomenon occurred which means that the population members grew during the GP execution without significant improvement of fitness (MAE) value. As a result the first and the third symbolic expression are much more complex than the second symbolic expression. It can also be noticed that the symbolic expressions used for gas turbine shaft torque estimation $X_1$, $X_3$, $X_4$, $X_5$, $X_8$, $X_9$, $X_{11}$ and $X_{12}$ are the most influential parameters in GT shaft torque estimation. In other words, input variables such as ship speed, gas generator rate of revolutions, starboard propeller torque, port propeller torque, GT turbocompressor outlet air temperature, HP turbine exit pressure, GT turbocompressor outlet air pressure and GT exhaust gas pressure are the most influential to the gas turbine shaft torque estimation. The real and estimated variation obtained using symbolic expressions in Tab. 5 of gas turbine shaft torque versus the ship speed is shown in Fig. 4.

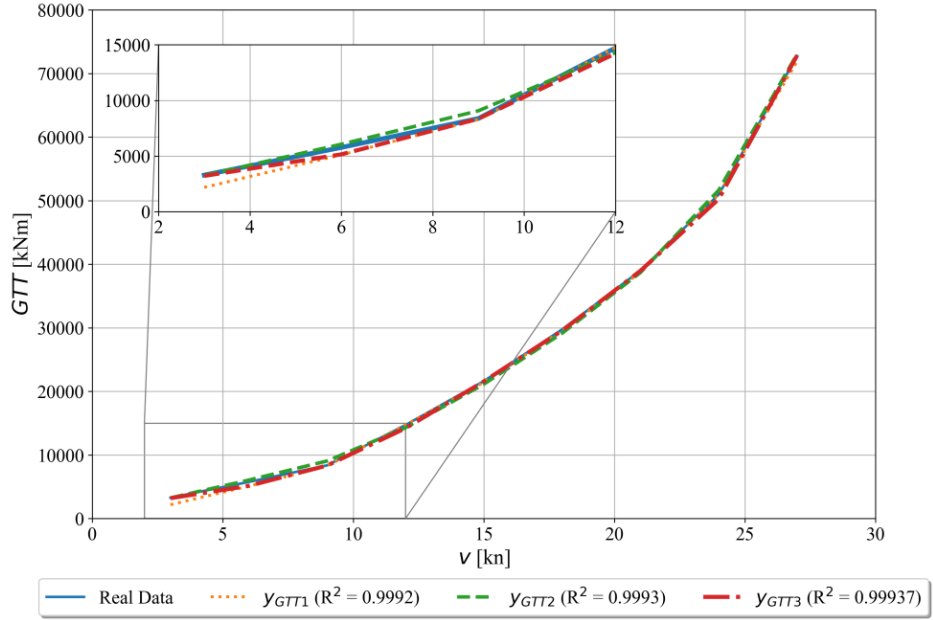

Figure 4. The real and estimated variation obtained using symbolic expressions of gas turbine shaft torque versus the ship speed. **Source:** Authors

As seen from Fig. 4, the real data (data from the dataset) shows that gas turbine shaft torque influences the ship speed. As the gas turbine shaft torque increases the ship speed also increases and the general curve trend is non-linear. The results obtained using three symbolic expressions from Tab. 5 have almost the same trend as the real data. The estimation of gas turbine shaft torque is very accurate with extremely deviation at lower ship speeds (from 3 to 6 kn) for all three symbolic expressions.

From conducted analysis it can be concluded that second symbolic expressions in Tab. 5 is the best symbolic expression to use in terms of equation length and accuracy. The third symbolic expression has a slightly higher accuracy but its much larger than the second symbolic expression.

### 3.2 Results Obtained for Fuel Flow

The training and testing portions of the dataset had 17 input variables and fuel flow values as the output variable. The 17 input variables are labeled from $X_0, \ldots, X_{16}$ while the output variable is labeled as *y*. The physical variables of the dataset with corresponding variable type used in GP algorithm to obtain the symbolic expressions for fuel flow estimation are shown in Tab. 2. The parameters used to obtain symbolic expressions for fuel flow estimation are shown in Tab. 6.

Table 6. The range of parameters used in GP to obtain symbolic expression for fuel flow estimation

| Parameter | Lower Bound | Upper Bound |
|---|---|---|
| Population size | 50 | 500 |
| Number of Generations | 100 | 300 |
| Tournament size | 5 | 50 |
| Tree depth | (3,7) | (6,12) |
| Crossover Coefficient | 0.7 | 1 |
| Subtree mutation coefficient | 0.01 | 0.1 |
| Hoist mutation coefficient | 0.01 | 0.1 |
| Point mutation coefficient | 0.01 | 0.1 |
| Stopping criteria | 1.00E-06 | 0.001 |
| Maximum Samples (%) | 90 | 100 |
| Constant range | -0.1 | 0.1 |
| Parsimony coefficient | 0.0001 | 0.01 |

**Source:** Authors

Three best symbolic expressions for fuel flow estimation with GP parameters used and resulting $R^2$ scores are shown in Tab 7.

Table 7. The symbolic expressions for fuel flow estimation with corresponding GP parameters and $R^2$ scores.

| GP Parameters | Symbolic Expression | $R^2$ score |
|---|---|---|
| [253, 292, 39, (3, 8), 0.7578, 0.0888, 0.0458, 0.049, 0.000725, 0.905, (-0.00878, 0.0152), 0.0036] | $y_{mf1} = \sqrt{X_7}\sqrt{\sqrt{\cos(\sin(\log(X_3)))}\log(X_{10})\log(X_{13})}$ | 0.995495 |
| [380, 131, 8, (3, 11), 0.759, 0.051, 0.0109, 0.093, 0.000856, 0.93, (-0.097, 0.0998), 0.0006] | $y_{mf2} = \sin(\log(\frac{X_9}{X_{16}} + \max((X_2 + X_6 + \tan(\log(X_6 \tan(\sin(X_{13}))))^{\frac{1}{2}}, X_6\tan(\sqrt{\sin(\log(X_7))})))) /\tan(\sqrt{\frac{\sin(\log(X_1 + X_4 + \min(X_6, X_7)))}{\sin(X_{13})}})$ | 0.996465 |
| [363, 122, 26, (5, 7), 0.758, 0.068, 0.057, 0.0214, 0.000103, 0.994, (-0.0581, 0.057), 0.0002] | $y_{mf3} = \left(\frac{1}{X_{15}}\right)\max\left(\sin(\cos(\sin(\sin(X_3)))), \frac{X_6}{X_8}\right) \min(\log(|\frac{X_{10}}{X_{15}}|), \sin(\sin(X_{13})))$ | 0.996487 |

**Source:** Authros

As seen from Tab. 7 the symbolic expressions for fuel flow estimation generated the $R^2$ score of $0.995495, 0.996465$ and $0.996487$, respectively. This symbolic expressions were obtained with randomly selected GP parameter values from ranges shown in Tab. 6. The population size in each generation consisted of 253, 380 and 363 population members evolved throughout 292, 131 and 122 generations. In each generation 39, 8 and 26 population members were randomly selected in each generation and compete to become parents for creation of next generation members. It should be noted that the lower tournament size value ensures higher diversity in the population. The randomly selected three depth range used for creation of initial population of naive formulas in each case was (3,8), (3,11) and (5,7). Each population member randomly chooses its maximum tree depth from specified range. The higher the tree depth value usually generates complex symbolic expressions which can be time consuming. The crossover coefficient in all three cases was around 0.758 and when compared to the other three mutation coefficient it can be concluded that crossover was the dominating variation operator in these GP analysis. The subtree mutation, hoist mutation and point mutation coefficient were randomly chosen and for the first case the values are 0.088, 0.0458 and 0.0.049; for the second case the values are 0.051, 0.0109 and 0.093 and for the third case the values are 0.068, 0.0057 and 0.0214, respectively. It can be noticed that these values in all three cases are very low when compared to the crossover coefficient values so their influence is significantly low. The stopping criteria coefficients in all three cases was randomly chosen value and are equal to 0.000725, 0.000856 and 0.000103, respectively. This criteria is responsible for stopping the execution of GP algorithm if the MAE value drops below the predefined stopping criteria value. However, in all three cases this criteria was never met and the GP algorithm execution stopped when the maximum number of generation was reached. The maximum number of training samples represents a fraction from training dataset used to evaluate each population member in each generation were randomly chosen values equal to 0.905, 0.994 and 0.994, respectively. The constant ranges in first, second and third case that were used to for development each population member and throughout the GP execution were randomly chosen ranges that are equal to (-0.00878, 0.0152), (-0.097, 0.0998) and (-0.0581, 0.057), respectively. The parsimony coefficients for first, second and third case were randomly chosen values that are equal to 0.036, 0.0006, and 0.0002. This coefficient is responsible for constant penalization of large programs by adjusting their fitness to be less favorable for selection. The larger values of this coefficient penalizes the population members more which can prevent the bloat phenomenon. As mentioned earlier, this phenomenon occurs when GP evolution is increasing the size of the population members without and significant increase in fitness value. However, in these analysis the values of parsimony coefficient were very low in order to enable the population member growth and the bloat phenomenon did not occur.

In all three symbolic expressions shown in Tab. 7 it can be noticed that these expressions consist of input variables $X_1, X_2, X_3, X_4, X_6, X_7, X_8, X_9, X_{10}, X_{13}, X_{15}$, and $X_{16}$. From Tab. 2 it can be noticed that input variables $X_5, X_{11}, X_{12}$ and $X_{14}$ have not any influence on fuel flow estimation. In other words the input variables starboard propeller torque ($T_s$), GT turbocompressor inlet air pressure ($P_1$), GT turbocompressor outlet air pressure ($P_2$) and turbine injection control (TIC) are omitted by GP in the aforementioned symbolic expressions. In Fig. 5 the performance of three symbolic expression given in Tab. 7 are shown versus the ship speed and compared to the fuel flow from the dataset.

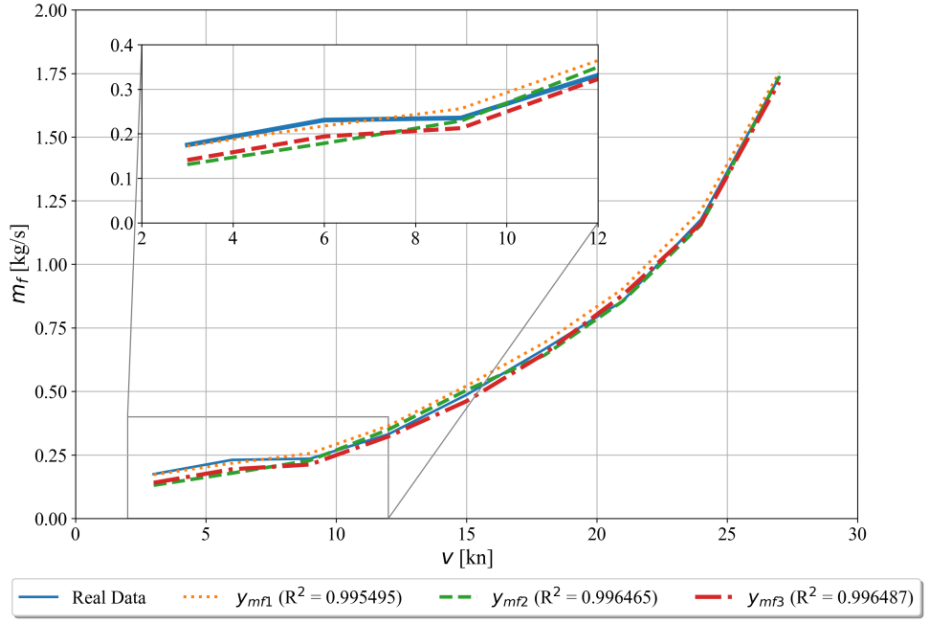

Figure 5. The variation of fuel flow versus the ship speed for real data and data obtained from symbolic expressions. **Source:** Authors

As seen from Fig. 5 the real data (data from the dataset) shows that as the ship speed increases the fuel flow also increases. All three symbolic expressions from Tab. 7 showed high accuracy in fuel flow estimation when compared to the real data with smaller deviations when sheep speed is in range form 3 to 6 kn.

From the following analysis it can be noticed that all three symbolic expressions shown in Tab. 7 can be used to estimate the fuel flow since the $R^2$ score for all three symbolic expressions is near 1.0. However, in terms of symbolic expression length the shortest symbolic expression when compared to the remaining two is the one with $R^2$ score of 0.995495. The other two symbolic expressions have slightly higher $R^2$ score (0.996465 and 0.996487). However, they are much more complex and longer than the first symbolic expression. So, from this it can be concluded that the first symbolic expression is good enough to be used for fuel flow estimation.

### 3.3 Discussion

The first analysis showed that all three equations are very accurate in gas turbine shaft torque estimation since the deviations from the real data are extremely small. All three symbolic expressions are obtained using small population sizes that are evolved throughout large number of generations (almost 1000 generations). In each generation in all three cases a small number (8, 20 and 23) were competing against each other to become the parents of the next generation. By pre-defining the small tournament size value ensures that only high quality population members in terms of fitness function values were competing against each other to become parents of the next generation. The dominating variation operator in all three symbolic expressions was crossover with coefficient value above 0.7. The mutation coefficients values when compared to the the crossover coefficient values are very small. The stopping criteria in all three case were extremely low values. Unfortunately in each case this criteria was never met so the maximum number of generations was utilized in order to terminate GP execution. So in each case the GP execution was terminated after the predefined maximum number of generations is achieved. The maximum number of training samples used to evaluate population members in each generation in all three cases was in range for 77 to 98.6 %. The highest constant range used for creation of population members was in first case from -77.601 up to 44.708 while the smallest range was used in third case from -3.56 up to 54.173, respectively. The parsimony coefficient value used in all three cases is very small in order to ensure the growth of the symbolic expression from generation through generation. However, the low parsimony coefficient value can cause the bloat phenomenon but in these analysis it did not occur.

As stated earlier the input values that ended up in all three symbolic expressions for gas turbine shaft torque estimation are $X_1, X_2, X_3, X_4, X_5, X_7, X_8, X_9, X_{11}, X_{12}$ and $X_{14}$. From the Tab. 2 it can be noticed that not all input variables are used for gas turbine shaft torque estimation. The lever position ($l_p$) high pressure turbine exit temperature ($T_{48}$), turbine injection control (TIC), GT turbocompressor decay state coefficient and GT turbine decay state coefficient are not used in symbolic expressions for gas turbine shaft torque estimation.

The second set of analysis showed that the best three symbolic expressions shown in Tab. 6 are very accurate in fuel flow estimation since the deviation from the real data are almost negligible. The GP parameters used used to obtain these symbolic expression are similar to those used to obtain symbolic expressions for gas turbine shaft torque estimation with smaller differences. The number of population members are in range from 200 to 300 while the maximum number of generations for all three cases is in range from 100 to 300. The tournament size range in all three cases is from 8 to 39 which is slightly larger than in the previous set of analyses. The crossover coefficient is the dominating variation operator when compared to the mutation coefficient values. The stopping criteria value in all three cases is very small as in the cases of symbolic expressions used for gas turbine shaft torque estimation. Again this criteria was never met and the GP execution in all three cases was terminated when the maximum number of generations is reached. The maximum number of training samples is higher than in previous set of analyses and is in range from 90.5 up to 99.4 %. The constant range used for creation of population members is extremely small when compared to the previous set of analyses. The parsimony coefficient values are extremely small as in the case of GP parameters used to obtain symbolic expressions for gas turbine shaft torque estimation.

The input parameters that ended up in all three symbolic expressions for fuel flow estimation are $X_1, X_2, X_3, X_4, X_6, X_7, X_8, X_9, X_{10}, X_{13}, X_{15}$ and $X_{16}$. When the variables used in all three symbolic expressions are compared with the list of variables shown in Tab. 2 it can be noticed that not all variables were used by GP to create all three symbolic expressions for fuel flow estimation. The lever position ($l_p$), starboard propeller torque ($T_s$), GT turbocompressor inlet air pressure ($P_1$), GT turbocompressor outlet air pressure ($P_2$) and turbine injection control (TIC) where are not used in these symbolic expressions.

Both analysis showed that all six symbolic expressions have very high accuracy in gas turbine shaft torque and fuel flow estimation. However, in terms of symbolic expression size the shortest symbolic expressions for gas turbine shaft torque estimation is the second symbolic expression. This symbolic expression has slightly lower accuracy in gas turbine shaft torque estimation ($R^2 = 0.999236$) when compared to the third symbolic expression ($R^2 = 0.999374$) which is far to long in terms of expression size. The shortest symbolic expression which can be used for fuel flow estimation is the first symbolic expressions shown in Tab. 7 with accuracy of $R^2 = 0.995495$. The remaining two symbolic expressions are larger than the first one with slightly higher accuracies ( $R^2 = 0.996465$ and $R^2 = 0.996487$ ). With these comparisons the far less complicated symbolic expressions can be used for gas turbine shaft torque estimation and fuel flow estimation without sacrificing the accuracy of the output values.

# 4 Conclusion

In this paper the publicly available condition based maintenance CODLAG dataset was used with GP algorithm in order to obtain symbolic expressions which could estimate the gas turbine shaft torque and fuel flow. From conducted investigation it can be concluded:

- the GP algorithm generated the symbolic expressions for gas turbine shaft torque estimation with high accuracy without the inclusion of lever position, high pressure turbine exit temperature, turbine injection control, GT turbocompressor decay state coefficient and GT turbine decay state coefficient,
- the GP algorithm generated the symbolic expressions for fuel flow estimation with high accuracy without the inclusion of lever position, starboard propeller torque, GT turbocompressor inlet air pressure, GT turbocompressor outlet air pressure and turbine injection control, and
- the symbolic expressions for gas turbine shaft torque estimation did not include the GT turbocompressor decay state coefficient and GT turbine decay state coefficient while the symbolic expressions for fuel flow estimation include two aforementioned coefficients.

## Acknowledgment

This research has been (partly) supported by the CEEPUS network CIII-HR-0108, European Regional Development Fund under the grant KK.01.1.1.01.0009 (DATACROSS), project CEKOM under the grant KK.01.2.2.03.0004, CEI project "COVIDAi" (305.6019-20) and University of Rijeka scientific grant uniri-tehnic-18-275-1447